\documentclass[letterpaper]{article} 
\usepackage{aaai25}  
\usepackage{times}  
\usepackage{helvet}  
\usepackage{courier}  
\usepackage[hyphens]{url}  
\usepackage{graphicx} 
\usepackage{natbib}  
\usepackage{caption} 
\usepackage{newfloat}
\usepackage{listings}
\DeclareCaptionStyle{ruled}{labelfont=normalfont,labelsep=colon,strut=off} 
\usepackage{subcaption}
\usepackage{amsmath}
\usepackage{amsfonts}
\usepackage{amsthm}
\usepackage{booktabs}
\usepackage{xspace}
\usepackage{amssymb}
\usepackage{todonotes}
\usepackage{pifont}
\usepackage{mathtools}
\usepackage{environ}
\usepackage{multirow}
\usepackage{pdfpages}
\usepackage{caption, array}
\usepackage{arydshln}
\usepackage{algpseudocode}  
\usepackage[ruled,vlined,linesnumbered]{algorithm2e}

\title{Trigger$^3$: Refining Query Correction via Adaptive Model Selector}
\author{
    Kepu Zhang,\textsuperscript{\rm 1} \ 
    Zhongxiang Sun,\textsuperscript{\rm 1} \ 
    Xiao Zhang,\textsuperscript{\rm 1,}\thanks{Corresponding author: Xiao Zhang (zhangx89@ruc.edu.cn).
    Work partially done at Engineering Research Center of Next-Generation Intelligent Search and Recommendation, Ministry of Education. 
    Work done when Kepu Zhang and Zhongxiang Sun were interns at Kuaishou.}\ 
    Xiaoxue Zang,\textsuperscript{\rm 2} \ \\
    Kai Zheng,\textsuperscript{\rm 2} \
    Yang Song,\textsuperscript{\rm 2} \
     Jun Xu\textsuperscript{\rm 1} 
}
\affiliations{
    \textsuperscript{\rm 1} Gaoling School of Artificial Intelligence, Renmin University of China \\
    \textsuperscript{\rm 2} Kuaishou Technology Co., Ltd.\\
    \{kepuzhang, sunzhongxiang, zhangx89\}@ruc.edu.cn\\
    \{zangxiaoxue,zhengkai,yangsong\}@kuaishou.com, junxu@ruc.edu.cn
}

\usepackage{bibentry}

\begin{document}

\maketitle

\begin{abstract}
In search scenarios, user experience can be hindered by erroneous queries due to typos, voice errors, or knowledge gaps. Therefore, query correction is crucial for search engines.
Current correction models, usually small models trained on specific data, often struggle with queries beyond their training scope or those requiring contextual understanding. While the advent of Large Language Models (LLMs) offers a potential solution, they are still limited by their pre-training data and inference cost, particularly for complex queries, making them not always effective for query correction.
To tackle these, we propose $\mathrm{Trigger}^3$, 
a large-small model collaboration framework that integrates the traditional correction model and LLM for query correction, 
capable of adaptively choosing the appropriate correction method based on the query and the correction results from the traditional correction model and LLM.
$\mathrm{Trigger}^3$ first employs a correction trigger to filter out correct queries. Incorrect queries are then corrected by the traditional correction model. If this fails, an LLM trigger is activated to call the LLM for correction. Finally, for queries that no model can correct, a fallback trigger decides to return the original query.
Extensive experiments demonstrate $\mathrm{Trigger}^3$ outperforms correction baselines while maintaining efficiency.
\end{abstract}

\section{Introduction}\label{sec:intro}
In online search scenarios, users may input incorrect queries due to insufficient knowledge, voice input, etc., resulting in errors such as typos, missing characters, homophones, and similar shapes~\cite{ye2023improving,pande2022learning}. If we do not correct the queries and use the original queries for searching, the results may significantly deviate from the user's needs. Therefore, to improve the user's search experience, search engines must implement query correction services that automatically detect and correct errors in queries.

\begin{figure}[t]
    \centering
        \begin{subfigure}{0.48\linewidth}
        \centering
    \includegraphics[width=\textwidth]{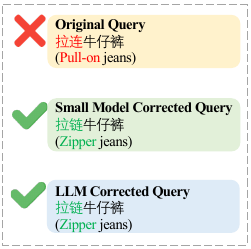}
        \subcaption{Query correction that requires common sense.}
    \end{subfigure}
        \begin{subfigure}{0.48\linewidth}
        \centering
    \includegraphics[width=\textwidth]{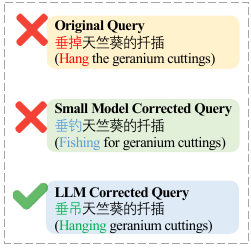}
        \subcaption{Query correction that requires context understanding.}
    \end{subfigure}
    \begin{subfigure}{0.995\linewidth}
        \centering
    \includegraphics[width=\textwidth]{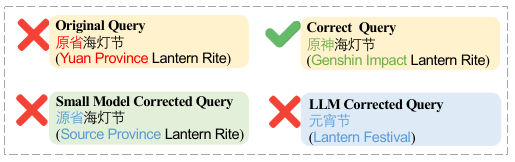}
        \subcaption{Query correction that requires specific domain knowledge.}
    \end{subfigure}
    \caption{Examples of query correction, where the red characters are the original errors, the blue characters are the results of corrected but incorrect, and the green characters are the correct result. The small model is traditional correction model GECToR and the LLM is Qwen1.5-7B-Chat.}
\label{fig:intro}
\end{figure}

In the field of query correction, the existing mainstream correction models can be divided into Seq2Seq and Seq2Edit methods. The Seq2Seq model~\cite{shao2024cpt,xue2021mt5} treats the correction task as a machine translation task, that is, translating the incorrect query into the correct query; the Seq2Edit model~\cite{zhang2022mucgec} treats the correction task as a sequence labeling task, correcting errors by marking insertions, deletions, etc. In this paper, \emph{we refer to these two types of traditional correction models as small models.} Nowadays, Large Language Models (LLMs) have demonstrated robust semantic comprehension in numerous tasks~\cite{brown2020language,ouyang2022training}, making them a viable option for query correction. When using the small model and LLM for query correction, we anticipate the following three observations:

\begin{itemize}
    \item 
Some queries are related to grammatical errors, which can be corrected based on common sense, where common sense refers to the knowledge that is easily included in the small model or LLM pre-training data, a capability possessed by both small and large models~\cite{ding2024hybrid}. 
For instance, as shown in Figure~\ref{fig:intro} (a), a user mistakenly inputs ``pull-on'' instead of ``zipper'' in the query due to a grammatical error. ``pull-on'' is not common in Chinese, while the correct ``zipper'' is very common, thus both models can correct it. Therefore, \emph{both small models and LLMs are capable of correcting errors in queries that can be addressed with common sense}.
\item 
Some queries necessitate a comprehensive understanding of query context, which may pose challenges for small models. For example, as depicted in Figure~\ref{fig:intro} (b), a user incorrectly inputs ``Hanging geranium cuttings'' as ``Hang the geranium cuttings''. GECToR corrects it to ``Fishing for geranium cuttings''. The words ``fishing'', ``hanging'', and ``hang'' are all grammatically correct in Chinese with similar pronunciations but vastly different meanings. Therefore, \emph{small models cannot correct errors in queries that require strong contextual semantic understanding, while LLMs can}.
\item 
As user queries may cover various aspects, there are certain queries that even the LLM might struggle to handle. These could be queries related to real-time news or specific domains. For example, As depicted in Figure~\ref{fig:intro} (c), within the gaming field, a user incorrectly inputs ``Genshin Impact Lantern Rite'' as ``Yuan Province Lantern Rite''. The small model corrects it to ``Source Province Lantern Rite'', while the LLM corrects it to ``Lantern Festival''. Both the small model and LLM, lacking knowledge in this specific domain, provide incorrect corrections. We observe that the corrected queries by the models might completely deviate from the user's original input. Using these deviated results as the final queries can severely affect the user search experience. Therefore, \emph{neither small models nor LLMs can correct errors in queries related to specific domains or real-time news}.
\end{itemize}

From these observations, we can learn that neither small models nor LLMs are universally effective in query correction tasks. Moreover, in terms of correction costs, the expenditure for small models is typically less than that for LLMs~\cite{ramirez2024optimising}. Therefore, the crucial issues when relying on small models and LLMs for query correction tasks are: \emph{when to employ either model and which one to choose for query correction, the small model or the LLM?} 
This is essentially a model selection problem for large-small model collaboration tasks, aimed at improving model performance and efficiency to enhance the trustworthiness \cite{liu2023trustworthy} and controllability \cite{shen2024control} of LLM-powered systems. 

To address the aforementioned issues, in this paper, we propose a novel model selector framework for query correction, named \textbf{$\mathrm{Trigger}^3$}, to adaptively integrate the small model and LLM for query correction. 
$\mathrm{Trigger}^3$ mainly consists of three parts:  Correction Trigger (CT), LLM Trigger (LT) and Fallback Trigger (FT). 

For when to employ models for correction: 
The CT selects incorrect queries for subsequent correction.
The FT conducts a review after the correction by both models, returning the original query for those that are difficult for both models to correct. 
For which model to choose: 
The LT selects queries that are difficult for the small model to correct but can be corrected by the LLM to the LLM for correction. In cases both models can correct, the small model's corrections are taken as final queries. 
Through the three modules, we not only leverage the correction capabilities of both models but also consider their limits, leading to enhanced correction performance and efficiency.

To validate the effectiveness and efficiency and of the proposed $\mathrm{Trigger}^3$ framework, we conduct experiments on two query correction datasets, using three small models and two LLMs. The results consistently demonstrate that $\mathrm{Trigger}^3$ achieves optimal performance and high efficiency. We summarize our contributions as follows:
\begin{itemize}
    \item We propose $\mathrm{Trigger}^3$, a novel large-small model collaboration framework that adaptively completes query correction by considering feedback from both the small model and LLM, which  is model-agnostic.
    \item We explore the combination of the small models and LLMs in the field of query correction, providing solutions for applying LLMs in query correction and how small models and LLMs can better collaborate.
    \item We conduct extensive experiments on both commercial and public datasets, showing that $\mathrm{Trigger}^3$ achieves superior performance while maintaining high efficiency.
\end{itemize}

\section{Related Work}
\subsection{Query Correction in Search Engines}
With the rise of neural networks, the current query correction models are mainly divided into two types: Seq2Edit and Seq2Seq. Seq2Edit models~\cite{zhang2022mucgec,awasthi2019parallel,liang2020bert} treat correction as a sequence tagging problem, completing the correction through editing operations such as insertion and deletion. Seq2Seq models~\cite{shao2024cpt,zhang2021mengzi,zhao2020maskgec} view the correction task as a translation task, translating the incorrect query into the correct one. They can achieve decent correction performance to a certain extent, but due to insufficient knowledge or weaker semantic understanding, they struggle to handle some queries.

Recently, some work has explored the application of Large Language Models (LLMs) in the correction field. By designing prompts and conducting a comprehensive evaluation of ChatGPT's performance on the correction task through in-context learning, ~\cite{fang2023chatgpt,li2023effectiveness,davis2024prompting,coyne2023analysis} find that LLMs tend to over-correct, and there is still a significant gap between LLM and small models trained on specific correction datasets. \cite{fan2023grammargpt} confirms that fine-tuning can enhance LLM's ability in text correction.
In this paper, we consider the correction feedback of small models and LLMs to jointly complete the query correction task.

\subsection{Model Selection of Language Models}
Model selection has long been a fundamental problem in machine learning~\cite{ding2018model,Zhang2019Survey,Zhang2020Hypothesis}.
Considering the high cost of LLMs, recent work has explored how to balance performance and efficiency. Their methods are mainly divided into two categories. 
The first category selects small and large models through a routing approach, mainly by predicting the accuracy of the small model's responses~\cite{lu2023routing,ding2024hybrid} to determine the invocation of the large model.
The second category adopts a cascading approach to decide whether to invoke the larger model after the execution of the smaller one. \cite{madaan2023automix} uses few-shot learning within the small model to verify its answers. \cite{yue2023large} judges based on the consistency of multiple answer samples obtained by the small model. In code-driven QA tasks, \cite{zhang2023ecoassistant} introduces an automatic code executor to decide based on the generated code execution. Most recently, ~\cite{ramirez2024optimising} makes decisions based on the uncertainty of the small model's output.

Unlike the above methods, we consider the specificity of query correction, which does not necessarily require an answer. Firstly, if the query is already correct, there's no need for correction. Secondly, both small and large models may not always provide accurate corrections. Hence, we designed the CT and FT to address these considerations.

\section{Trigger$^3$: The Proposed Framework}\label{sec:method}

\subsection{Task Formalization}
In the query correction task, we are given a set of data $\mathcal{D}=\{(x_i,y_i )\}_{i=1}^{|D|}$, where $|D|$ indicates the total number of data, each of these data samples contains: $x_i$ represents the $i$-th original query, $y_i$ represents the correct query corresponding to $x_i$.
The goal of the query correction task is to learn the function from the original query to the target query. Here query $x_i$ and $y_i$ may not be the same length.

\begin{figure*}
    \centering
\includegraphics[width=0.98\linewidth]{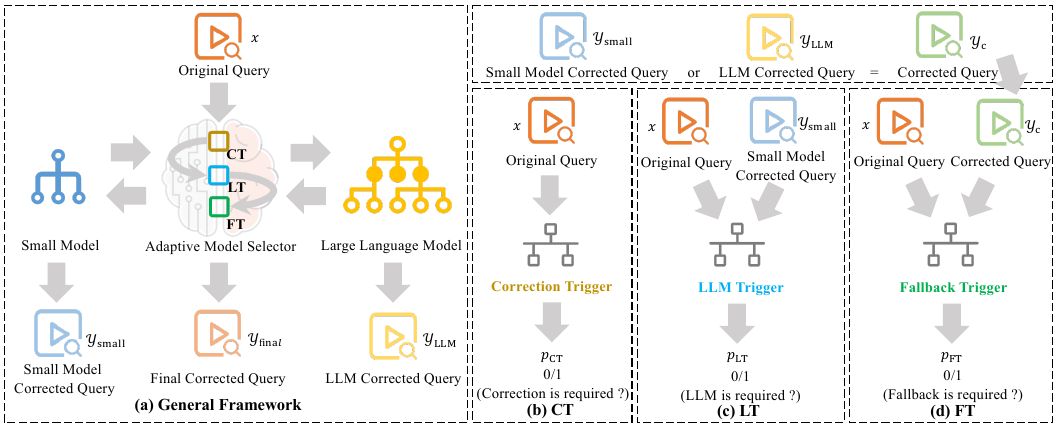}
    \caption{
    The architecture of the proposed framework $\mathrm{Trigger}^3$. (a) The general framework of $\mathrm{Trigger}^3$. (b) The Illustration of Correction Trigger (CT). (c) The Illustration of LLM Trigger (LT). (d) The Illustration of Fallback Trigger (FT).}
    \label{fig:model_graph}
\end{figure*}

\subsection{General Framework}
The large-small model collaboration framework of $\mathrm{Trigger}^3$ is shown in Figure~\ref{fig:model_graph}. The input is the original query, and after interacting with the adaptive model selector, the small model and LLM, the output is the final corrected query. 

\begin{algorithm}[t]
	\caption{Process flow of $\mathrm{Trigger}^3$.}
	\label{alg:qc process}
\textbf{Input:} Original query $x$ and $\mathrm{Trigger}^3$'s models.\\
\textbf{Output:} Final corrected query $y_{\mathrm{final}}$.\\
	$p_{\mathrm{CT}} \leftarrow f_{\mathrm{CT}}(x)$ \Comment{Correction Trigger}\label{line:ct}
 
	\If{$p_\mathrm{CT}=1$}{
        $y_{\mathrm{small}} \leftarrow f_{\mathrm{small}}(x)$
        
        $p_{\mathrm{LT}} \leftarrow f_{\mathrm{LT}}(x,y_{\mathrm{small}})$  
        \Comment{LLM Trigger}\label{line:lt}

        \If{$p_\mathrm{LT}=1$}{
        $y_{\mathrm{LLM}} \leftarrow f_{\mathrm{LLM}}(x,y_{\mathrm{small}})$

        $y_{\mathrm{c}}=y_{\mathrm{LLM}}$
        }
        \Else{
        $y_{\mathrm{c}}=y_{\mathrm{small}}$
        }

        $p_{\mathrm{FT}} \leftarrow f_{\mathrm{FT}}(x,y_{\mathrm{c}})$  \Comment{Fallback Trigger}\label{line:ft}

        \If{$p_\mathrm{FT}=1$}{
            $y_{\mathrm{final}}=x$
        }
        \Else{
            $y_{\mathrm{final}}=y_{\mathrm{c}}$ 
        }        
    }
    \Else{
    $y_{\mathrm{final}}=x$
    }

\end{algorithm}
\setlength{\textfloatsep}{0.1pt}

The adaptive model selector consists of 
1) \textbf{Correction Trigger (CT)} $f_{\mathrm{CT}}$ that decides whether the original query needs to be corrected,  
2) \textbf{LLM Trigger (LT)} $f_{\mathrm{LT}}$ that analyzes if the LLM is needed for query correction, and 
3) \textbf{Fallback Trigger (FT)} $f_{\mathrm{FT}}$ that checks 
whether the original query needs to be returned.
The details are covered in the next three sections.

As shown in Algorithm~\ref{alg:qc process}, a query $x$ is corrected following the process below:

The query will first go through the CT (line~\ref{line:ct}), which will determine whether it needs to be corrected based on its correctness. 
If the CT determines that the query needs to be corrected, it passes the query to the small model. This model is designed to handle common and simple errors and is more efficient compared to the LLM. We denote it as $f_{\mathrm{small}}$. $f_{\mathrm{small}}$ takes original query $x$ as input and outputs its corrected query $y_\mathrm{small}$:
\begin{equation}
\label{eq:small}
   y_{\mathrm{small}}=f_{\mathrm{small}}(x;\theta_{\mathrm{small}}),
\end{equation}
where $\theta_{\mathrm{small}}$ is the learnable parameters in small model. 
After being corrected by the small model, the query corrected by the small model and the original query will go through the LT (line~\ref{line:lt}) to determine whether the LLM is needed for correction.

If the LT determines that the query cannot be corrected by the small model, but can be corrected by the LLM, the query is passed to the LLM. This model is more powerful and can handle more complex errors, but it is more resource-intensive. We denote it as $f_{\mathrm{LLM}}$. 
$f_{\mathrm{LLM}}$ takes $(x,y_{\mathrm{small}})$ as input and outputs the its corrected query $y_{\mathrm{LLM}}$:
\begin{equation}
\label{eq:llm}
   y_{\mathrm{LLM}}=f_{\mathrm{LLM}}(x, y_{\mathrm{small}};\theta_{\mathrm{LLM}}),
\end{equation}
where $\theta_{\mathrm{LLM}}$ is parameters in $f_{\mathrm{LLM}}$.

Finally, the FT (line~\ref{line:ft}) will determine whether to return the original query as the final query output based on the corrected query and the original query.
That is, the final corrected query may use the corrections from the small model, the LLM, or it may remain the original query:
\begin{equation}
   y_{\mathrm{final}}=x\; \mathrm{or} \; y_{\mathrm{small}} \; \mathrm{or} \; y_{\mathrm{LLM}}
\end{equation}

\subsection{The First Trigger: Correction Trigger}
To improve efficiency, we first judge the correctness of the original query. If the query itself is correct, there is no need to use the small model and the LLM for correction. 

We use the Correction Trigger (CT) to achieve the above goal. Given the initial query $x$, CT is a scoring function that indicates the probability of the query being incorrect:
\begin{equation}
\label{eq:ct}
\begin{aligned}
   p_{\mathrm{CT}} =P(\text{Incorrect}|x) 
   =f_{\mathrm{CT}}(x;\theta_{\mathrm{CT}}),
\end{aligned}
\end{equation}
where $P(\text{Incorrect}|x)$ is the probability that the query $x$ is incorrect. If $p_{\mathrm{CT}}$ is above a certain threshold, we can conclude that the query is incorrect and correction is needed. 

We use the representation of the [CLS] token in BERT~\cite{devlin2019bert} to get the score $p_{\mathrm{CT}}$.

\subsection{The Second Trigger: LLM Trigger}\label{sec:LT}
After the small model's correction, we use a LLM Trigger (LT) to decide whether to invoke the Large Language Model (LLM). Considering that the LLM may not be able to solve the problem either, we hope to use LT to identify the queries that the small model cannot correct but the LLM can.
Given the pair of the original query and the query preliminarily rewritten by the small model $(x,y_{\mathrm{small}})$, LT is a scoring function that indicates the probability of calling LLM:
\begin{equation}
\label{eq:lt}
\begin{aligned}
       p_{\mathrm{LT}}&=P(\text{Invoke} \; \text{LLM}|x, y_{\mathrm{small}})\\
   &=f_{\mathrm{LT}}(x,y_{\mathrm{small}};\theta_{\mathrm{LT}}),
\end{aligned}
\end{equation}
where $y_{\mathrm{small}}$ is the output of the small model. We use the [SEP] token to separate $x$ and $y_{\mathrm{small}}$, and take the representation of the [CLS] token to get the score $p_{\mathrm{LT}}$.

\subsection{The Third Trigger: Fallback Trigger}\label{sec:f trigger}
Considering that both small and large models may not be able to correct some queries such as real-time news queries or domain-specific queries, which, if modified, may seriously damage the user search experience, as shown in Figure~\ref{fig:intro} (b), it is better to use the original query. 
This operation is inspired by the research about LLM's refusal to answer~\cite{chen2024benchmarking} and LLM security~\cite{zheng2024prompt,sun2023safety}.

After the small model or LLM correction, we can review the rewrite and choose whether to return the original query based on the corrected query and the original query.
Given the pair of the original query and corrected query, $p_{\mathrm{FT}}$ is used to indicate the probability of returning the original query:
\begin{equation}
\label{eq:ft}
\begin{aligned}
       p_{\mathrm{FT}}&=P(\text{Return} \; x|x, y_{\mathrm{c}})\\
   &=f_{\mathrm{FT}}(x, y_{\mathrm{c}};\theta_{\mathrm{FT}}),
\end{aligned}
\end{equation}
where $y_\mathrm{c}$ is either $y_{\mathrm{small}}$ or $y_{\mathrm{LLM}}$, which can be known according to Algorithm~\ref{alg:qc process}.
We use the [SEP] token to separate $x$ and $y_{\mathrm{c}}$, and take the representation of the [CLS] token to get the score $p_{\mathrm{FT}}$.

\subsection{Model Training in $\mathrm{Trigger}^3$}\label{sec:model training}

In $\mathrm{Trigger}^3$, for the three modules, we use the widely used binary cross-entropy loss~\cite{devlin2019bert} as the objective function:
\begin{multline}
\label{eq:loss_xt}
    \mathcal{L}_{\mathrm{XT}} = - \frac{1}{|\mathcal{D}_{\mathrm{XT}}|} \sum_{\mathcal{D}_{\mathrm{XT}}} y_{\mathrm{XT}}\mathrm{log}(p_{\mathrm{XT}}) \\
    +(1-y_{\mathrm{XT}})\mathrm{log}(1-p_{\mathrm{XT}}),
\end{multline}
where $\mathrm{XT}\in\{\mathrm{CT},\mathrm{LT},\mathrm{FT}\}$, $y_{\mathrm{XT}}
$ is the label and $p_{\mathrm{XT}}$ is the prediction score.

For $\mathcal{D}_{\mathrm{CT}}$, we take the wrong query in the training dataset as the positive sample and the correct query as the negative sample.

Before introducing the dataset construction for $\mathcal{D}_{\mathrm{LT}}$ and $\mathcal{D}_{\mathrm{FT}}$, we first introduce a few character-edit-based indicators that will be used later: True positive (TP) indicates whether the model has correct edits, False positive (FP) indicates whether the model's edits have changed the correct characters into the wrong ones, and False negative (FN) indicates whether the model's edits have missed any necessary changes for the correct query. For the small model's editing indicators, we represent them as $\mathrm{TP}_\mathrm{S}$, $\mathrm{FP}_\mathrm{S}$, $\mathrm{FN}_\mathrm{S}$. For the LLM's editing indicators, we represent them as $\mathrm{TP}_\mathrm{L}$, $\mathrm{FP}_\mathrm{L}$, $\mathrm{FN}_\mathrm{L}$.

For $\mathcal{D}_{\mathrm{LT}}$, we use the queries that \textbf{small model can't correct, but LLM can} as the positive samples.
Specifically, a query is determined to be a positive sample for LT as long as it meets any of the following three points:
1) The small model does not have correct edits, but the LLM does.
2) The small model has incorrect edits, but the LLM does not.
3) The small model has missed necessary edits, but the LLM does not, i.e., the LLM has completely corrected this query.
This can be represented as 
\begin{equation}
\label{eq:lt train data}
\begin{aligned}
   &(\mathrm{TP}_\mathrm{S}<0 \quad \text{and} \quad \mathrm{TP}_\mathrm{L}>0) \\
   \text{or}\; &  (\mathrm{FP}_\mathrm{S}>0 \quad  \text{and} \quad \mathrm{FP}_\mathrm{L}<0) \\
    \text{or}\; &  (\mathrm{FN}_\mathrm{S}>0 \quad  \text{and} \quad \mathrm{FN}_\mathrm{L}<0). \notag
\end{aligned}
\end{equation}
Negative samples are then sampled in the same quantity as positive samples, excluding all positive samples from the training dataset.

For $\mathcal{D}_{\mathrm{FT}}$, we use the queries that \textbf{both small model and LLM cannot correct} as the positive samples.
Specifically, a query is determined to be a positive sample for FT if the editing of the rewritten query does not have a correct edit. We consider that both the small model and LLM do not have a correct edit, specifically represented as 
\begin{equation}
   \mathrm{TP}_\mathrm{S}<0 \quad \text{and} \quad \mathrm{TP}_\mathrm{L}<0. \notag
\end{equation}
Negative samples are then sampled in the training set, excluding all positive samples, with the same number of positive samples.

The training of the LLMs and the small models can be found in Section~\ref{sec:implementation details}.

\section{Experiments}\label{sec:experiments}

\begin{table}[t]
    \centering
        \begin{tabular}{l c c c}
            \toprule
            \textbf{Train}  & \textbf{Avg len}  & \textbf{\#Query}  &  \textbf{Error Rate}   \\
            \hline
            \textbf{Commercial}  & 9.43  & 1,444,213    & 97.8\% \\
            \textbf{QQ} & 9.81  & 111,703    & 79.1\% \\
            \bottomrule
            \textbf{Valid}  & \textbf{Avg len}  & \textbf{\#Query}  &  \textbf{Error Rate}   \\
            \hline
            \textbf{Commercial}  & 9.41  & 14,737    & 97.8\% \\
            \textbf{QQ} & 9.78  & 12,412    & 75.1\% \\
            \bottomrule
            \textbf{Test}  & \textbf{Avg len}  & \textbf{\#Query}  &  \textbf{Error Rate}   \\
            \hline
            \textbf{Commercial}  & 9.43  & 14,737    & 97.8\% \\
            \textbf{QQ} & 9.79  & 13,791    & 74.7\% \\
            \bottomrule
        
        \end{tabular}
    
    \caption{Statistics of the used query correction datasets. \textbf{Avg len} is the average length of the original query, \textbf{\#Query} denotes the number of the queries and \textbf{Error Rate} denotes the percentage of the incorrect queries.}
    \label{tab:dataset_statistics}
\end{table}

\begin{table*}[!ht]

\centering
\resizebox{\linewidth}{!}{
\begin{tabular}{ll cccccc cccccc}
\toprule 

\multirow{3}{*}{\textbf{Category}}&
\multirow{3}{*}{\textbf{Model}} & \multicolumn{6}{c}{\textbf{Commercial}} & \multicolumn{6}{c}{\textbf{QQ}}                                                                      \\
\cmidrule(lr){3-8}\cmidrule(lr){9-14}&
& \multicolumn{3}{c}{\textbf{Character-level}} & \multicolumn{3}{c}{\textbf{Word-level}}            & \multicolumn{3}{c}{\textbf{Character-level}} & \multicolumn{3}{c}{\textbf{Word-level}}                                                           \\
\cmidrule(lr){3-5}\cmidrule(lr){6-8}\cmidrule(lr){9-11}\cmidrule(lr){12-14}

  && \textbf{P}& \textbf{R}& \textbf{F$_{0.5}$} &  \textbf{P}    &  \textbf{R} & \textbf{F$_{0.5}$} & \textbf{P}& \textbf{R}& \textbf{F$_{0.5}$} &  \textbf{P}    &  \textbf{R} & \textbf{F$_{0.5}$} \\
  
\hline
\multirow{3}{*}{Individual} &
\textbf{GECToR} (Small Model) & 59.59 & \textbf{76.30} & 62.32 & 58.68 & \underline{68.71} & 60.44  & 39.96 & 46.10 & 41.05 & 44.59 & 43.69 & 44.41   \\
&Single (LLM) & 45.47 & 42.87 & 44.92 & 45.57 & 40.96 & 44.56   & 41.57 & 40.50 & 41.35 & 43.93 & 37.78 & 42.55\\
&Cascading (LLM) & \underline{72.43} & 67.13 & \underline{71.30} & \underline{72.34} & 64.35 & \underline{70.59}  & \underline{51.84} & \underline{47.00} & \underline{50.79} & \underline{54.66} & \underline{44.72} & \underline{52.34}   \\
\cdashline{1-14}
\multirow{6}{*}{Combination} &
Random Routing & 53.16 & 59.29 & 54.28 & 52.54 & 54.56 & 52.93 & 40.32 & 42.97 & 40.82 & 44.00 & 40.53 & 43.26  \\
&Meta Routing  & 59.08 & 70.25 & 61.02 & 58.70 & 64.71 & 59.81 & 43.38 & 46.77 & 44.02 & 46.20 & 43.24 & 45.58 \\
&HybridLLM & 59.63 & 71.82 & 61.72 & 59.10 & 65.76 & 60.32  & 43.57 & 46.51 & 44.13 & 46.60 & 43.13 & 45.86 \\
&Random Cascading & 64.61 & 71.60 & 65.90 & 64.35 & 66.33 & 64.73 & 45.07 & 46.30 & 45.31 & 48.96 & 43.89 & 47.85 \\
&Margin Sampling & 66.56 & 71.41 & 67.48 & 66.29 & 66.54 & 66.34 & 47.25 & 46.57 & 47.11 & 51.24 & 44.70 & 49.79 \\
&$\mathrm{Trigger}^3$ (Ours) & \textbf{74.66}$^\dagger$  & \underline{74.33} & \textbf{74.60}$^\dagger$  & \textbf{74.79}$^\dagger$  & \textbf{71.33}$^\dagger$  & \textbf{74.07}$^\dagger$   &  \textbf{60.09}$^\dagger$  & \textbf{48.69}$^\dagger$  & \textbf{57.40}$^\dagger$  & \textbf{63.45}$^\dagger$  & \textbf{46.96}$^\dagger$  & \textbf{59.29}$^\dagger$  \\
\midrule
\multirow{3}{*}{Individual} &
\textbf{BART} (Small Model) & \underline{73.52} & \underline{71.99} & \underline{73.21} & \underline{73.91} & \underline{71.54} & \underline{73.42}  & 59.83 & 60.51 & 59.97 & 62.26 & \underline{62.11} & 62.23   \\
&Single (LLM) & 45.47 & 42.87 & 44.92 & 45.57 & 40.96 & 44.56   & 41.57 & 40.50 & 41.35 & 43.93 & 37.78 & 42.55\\
&Cascading (LLM) & 72.73 & 62.55 & 70.43 & 73.05 & 61.79 & 70.48 & 55.73 & 52.41 & 55.03 & 58.57 & 51.32 & 56.96   \\
\cdashline{1-14}
\multirow{6}{*}{Combination} &
Random Routing  & 59.64 & 57.24 & 59.14 & 60.14 & 56.16 & 59.30 & 50.55 & 50.13 & 50.47 & 53.56 & 49.64 & 52.73 \\
&Meta Routing  & 68.68 & 65.77 & 68.08 & 69.06 & 64.97 & 68.20 & 60.78 & \underline{60.52} & 60.73 & 63.87 & 60.72 & 63.21 \\
&HybridLLM  & 71.12 & 68.64 & 70.61 & 71.66 & 68.08 & 70.91 & \underline{61.82} & 60.44 & \underline{61.54} & \underline{64.92} & 60.65 & \underline{64.02} \\
&Random Cascading & 73.23 & 67.43 & 71.99 & 73.73 & 66.83 & 72.24 & 57.52 & 56.03 & 57.21 & 60.11 & 56.21 & 59.29\\
&Margin Sampling & 72.67 & 66.52 & 71.35 & 72.95 & 65.88 & 71.41 & 58.73 & 58.46 & 58.67 & 61.51 & 59.16 & 61.03\\
&$\mathrm{Trigger}^3$ (Ours)  & \textbf{76.57}$^\dagger$  & \textbf{72.07} & \textbf{75.63}$^\dagger$  & \textbf{76.86}$^\dagger$  & \textbf{71.57} & \textbf{75.74}$^\dagger$  & \textbf{66.31}$^\dagger$  & \textbf{61.43}$^\dagger$  & \textbf{65.27}$^\dagger$  & \textbf{68.59}$^\dagger$  & \textbf{62.17} & \textbf{67.20}$^\dagger$   \\
\midrule
\multirow{3}{*}{Individual} &
\textbf{mT5} (Small Model) & 67.42 & 59.04 & 65.56 & 68.44 & 58.02 & 66.06  & 54.71 & 52.01 & 54.15 & 56.61 & \underline{51.02} & 55.40   \\
&Single (LLM) & 45.47 & 42.87 & 44.92 & 45.57 & 40.96 & 44.56   & 41.57 & 40.50 & 41.35 & 43.93 & 37.78 & 42.55\\
&Cascade (LLM)& 66.18 & 54.90 & 63.57 & 66.90 & 53.79 & 63.79  & 50.48 & 49.02 & 50.18 & 52.86 & 46.56 & 51.46 \\
\cdashline{1-14}
\multirow{6}{*}{Combination} &
Random Routing  & 56.05 & 51.00 & 54.96 & 56.66 & 49.60 & 55.09  & 47.78 & 45.86 & 47.38 & 50.32 & 44.06 & 48.93 \\
&Meta Routing  & 64.08 & 57.62 & 62.67 & 64.98 & 56.51 & 63.09 & 57.84 & 52.14 & 56.60 & 60.87 & 50.71 & 58.53 \\
&HybridLLM & 66.71 & 58.44 & 64.87 & 67.52 & 57.21 & 65.17 & \underline{59.14} & \underline{52.22} & \underline{57.61} & \underline{62.00} & 50.90 & \underline{59.41} \\
&Random Cascading  & 66.96 & 57.17 & 64.74 & 67.94 & 56.10 & 65.19& 52.29 & 50.01 & 51.82 & 54.27 & 48.08 & 52.91 \\
&Margin Sampling & \underline{67.51} &\underline{59.06} & \underline{65.63} & \underline{68.54} & \underline{58.04} &\underline{ 66.14} & 55.26 & 51.95 & 54.57 & 57.05 & 50.74 & 55.66 \\
&$\mathrm{Trigger}^3$ (Ours) & \textbf{69.64}$^\dagger$  & \textbf{59.13} & \textbf{67.25}$^\dagger$  & \textbf{70.44}$^\dagger$  & \textbf{58.10} & \textbf{67.57}$^\dagger$  & \textbf{61.36}$^\dagger$  & \textbf{52.72}$^\dagger$  & \textbf{59.41}$^\dagger$  & \textbf{64.43}$^\dagger$  & \textbf{51.29} & \textbf{61.29}$^\dagger$ \\
\bottomrule
\end{tabular}
}
\caption {Performance comparisons between $\mathrm{Trigger}^3$ and the baselines when the LLM is Qwen1.5-7B-Chat. 
Single: directly using LLM for correction. Cascading: using smaller model rewrites as part of LLM prompts. The LLMs use 1,000 data for fine tuning, while the small model use full training data for training.
The boldface indicates the best performance, and the underline indicates the second performance. `$\dagger$' indicates that the improvements are significant (t-tests, $p\textrm{-value}< 0.05$).
}
\label{tab:main results qwen} 
\end{table*}

\subsection{Experimental Settings}
\subsubsection{Dataset}
We conduct query correction experiments on the following two datasets:

\noindent \textbf{Commercial}
is based on the user search logs from a popular short video platform in 2024.  
The construction process of Commercial dataset is as follows: 50\% of the data is obtained by rejecting samples from online correction logs with a rewriting confidence greater than 0.99. The remaining 50\% of the data is generated from high-quality online queries through methods such as homophone substitution, near-sound character replacement, adjacent character transposition, and random character addition or deletion.

\noindent \textbf{QQ}
is a publicly available search-related dataset, 
due to the lack of publicly available query correction datasets, we modify it as a query correction dataset. 
Following~\cite{ye2023improving}, we first use a language model to filter the queries, selecting those with a high probability of being correct. We then perform similar operations like Commercial dataset on these queries to construct a query correction dataset.

The statistics and the construction process of the datasets are shown in Table~\ref{tab:dataset_statistics}.

\subsubsection{Metrics}
Following~\cite{xu2022fcgec}, we use the widely used metrics character-level and word-level precision (P)/recall (R)/F-measure (F$_{0.5}$) from ChERRANT scorer~\cite{zhang2022mucgec} to evaluate the correction performance. 

\subsubsection{Baselines}
In order to verify the validity of $\mathrm{Trigger}^3$, we consider the following correction model as the small model:
GECToR, BART, mT5, which are short for GECToR-Chinese~\cite{zhang2022mucgec}, BART-Large~\cite{shao2024cpt} and mT5-Base~\cite{xue2021mt5}.
We consider the following LLM:
Qwen1.5-7B-Chat~\cite{qwen} and Baichuan2-7B-Chat~\cite{yang2023baichuan}. We improve LLM's correction performance by fine-tuning it and applying it for direct correction (Single) and using small model rewrites as part of LLM prompts for corrections (Cascading). 
The reasons for fine-tuning can be found in Appendix~\ref{appendix llm issues}.

We further combine the small model and LLM and compare $\mathrm{Trigger}^3$ to the following framework:
Random-Routing, Routing~\cite{lu2023routing,vsakota2024fly}, HybridLLM~\cite{ding2024hybrid}, Random-Cascading and Margin Sampling~\cite{ramirez2024optimising}. 
Specifically, we compare the correction performance of GECToR, BART, mT5, LLM itself and with $\mathrm{Trigger}^3$. Then, using these small models and LLMs, we further compare $\mathrm{Trigger}^3$ with the above frameworks. 
The details of the above baselines are presented in Appendix~\ref{appendix:baseline}.

\subsubsection{Implementation Details}\label{sec:implementation details}
Our code implementation is based on Huggingface Transformers~\cite{wolf2020transformers} in Pytorch.
The fine tuning cost of LLM is much higher than that of small models. Therefore, following~\cite{fan2023grammargpt}, for the fine tuning of LLM, we only used 1,000 pieces of data from the training dataset, while for the training of small models, we used all available training datasets.
We train the small model according to the parameters of the original paper.
For the fine tuning of LLMs, we use LoRA~\cite{hu2021lora} for efficient fine tuning. We utilize the Adam~\cite{kingma2014adam} optimizer, setting the initial learning rate to 5e-5, the batch size to 16, and applying a cosine learning rate schedule for 3 epochs.
For a fair comparison, all cascading strategies provide preliminary rewrites from small models to LLM, enhancing the LLM's correction performance.
For the auxiliary models used in $\mathrm{Trigger}^3$ and all frameworks, we select ten thousand queries from the training dataset to fine-tune BERT~\cite{devlin2019bert}.
All experiments are performed on NVIDIA V100 32GB GPUs. 
More details about the implementation can be found in Appendix~\ref{appendix:implementation} and \url{https://github.com/ke-01/Trigger3}.

\subsection{Main Results}\label{sec:main results}

We investigate the correction performance of our proposed $\mathrm{Trigger}^3$. As shown in Table~\ref{tab:main results qwen}, which presents the correction performance on two datasets, we can draw the following conclusions:

\noindent$\bullet$ \textbf{Overall Performance}. $\mathrm{Trigger}^3$ surpasses all base small models, LLMs and frameworks in F$_{0.5}$ while ensuring no decrease in recall rate. This demonstrates the effectiveness of our proposed $\mathrm{Trigger}^3$ in integrating the small model and LLM, taking into account the feedback from both when deciding whether to call the LLM and returning the original query strategy for queries that neither model corrects well.

\noindent$\bullet$ \textbf{Cascading vs. Routing}. We find that the cascade framework performs better overall in correction than the routing framework. This is mainly because, in the correction task, without the preliminary rewriting from the small model, direct correction by the LLM may result in over-correction, leading to poorer correction performance. This suggests that in the query correction task, the preliminary rewriting by the small model can serve as an implicit feature to help improve the LLM's correction performance.

\noindent$\bullet$ \textbf{Comparison of Different Small Models}. For different small models, we note that combining with the LLM improves the performance of Seq2Edit more significantly. This is mainly because the types of errors that Seq2Edit and Seq2Seq can correct are more complementary. 
This also reflects to some extent that the errors Seq2Seq and LLM can solve may be more alike. 
However, as the errors that the LLM and Seq2Seq small model can correct are different, this can also enhance the base model's correction performance.

We perform the experiment with similar conclusions when LLM is Baichuan2-7B-Chat in Appendix~\ref{appendix:baichuan results}.

\subsection{Ablation Study}

\begin{table}[t]
\centering
{
    \resizebox{\columnwidth}{!}{
\renewcommand{\arraystretch}{1.0}
\begin{tabular}{lcccc}
\toprule

\multirow{2}{*}{\textbf{Model}} & \multicolumn{2}{c}{\textbf{Commercial}} & \multicolumn{2}{c}{\textbf{QQ}} \\
\cmidrule(lr){2-3}\cmidrule(lr){4-5}
   & Char-F$_{0.5}$ & Word-F$_{0.5}$ & Char-F$_{0.5}$ & Word-F$_{0.5}$  \\
    \midrule
    GECToR     &62.32&60.44&41.05       &44.41\\
    LLM       &71.30&70.59&42.29     &45.48\\
    \cdashline{1-5}
    $+\mathrm{LT}$       &73.33&72.83&55.91  &57.91\\
    \ \ $+\mathrm{FT}$      &74.17&73.66&56.63      &58.49\\
    $\quad+\mathrm{CT}$     &\textbf{74.60}&\textbf{74.07}&\textbf{57.40}     &\textbf{59.29}  \\
    \midrule
    BART       &73.21&73.42&59.97    &62.23 \\
    LLM       &70.43&70.48&52.79     &55.38\\
    \cdashline{1-5}
    $+\mathrm{LT}$      &74.90&75.03  &64.52 &66.40 \\
    \ \ $+\mathrm{FT}$   &75.21&75.33  &64.67   &66.58     \\
    $\quad+\mathrm{CT}$   &\textbf{75.63}&\textbf{75.74}  &\textbf{65.27}     &\textbf{67.20}  \\
    \midrule
        mT5          &65.56&66.06&54.15  &55.40\\
    LLM       &63.57&63.79&50.18     &51.46\\
    \cdashline{1-5}
    
    $+\mathrm{LT}$   &66.69&67.07 &57.22 &59.03     \\
    \ \ $+\mathrm{FT}$  &67.11&67.44   &58.28 &60.10       \\
    $\quad+\mathrm{CT}$  &\textbf{67.25}&\textbf{67.57}  &\textbf{59.41} &\textbf{61.29}        \\
    \bottomrule
\end{tabular}
}
}
    \caption{Ablation studies of $\mathrm{Trigger}^3$ on Commercial and QQ datasets when the LLM is Qwen1.5-7B-Chat. The boldface indicates the best performance.}
    \label{tab:ablation}
\end{table}

\begin{figure}[t]
    \centering
        \includegraphics[width=0.9\linewidth]{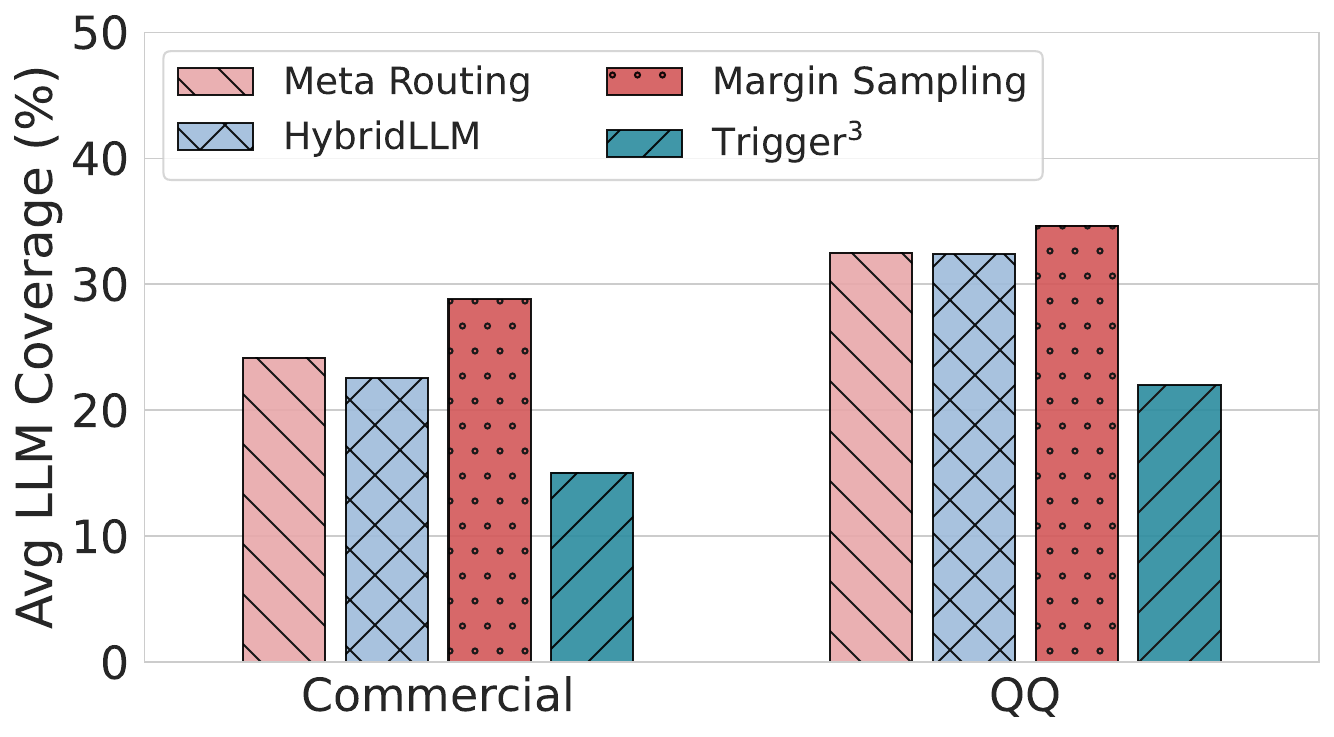}
    \caption{ Average LLM Coverage of $\mathrm{Trigger}^3$ and the three frameworks when the LLM is Qwen1.5-7B-Chat. The lower the bar, the better.
    }
\label{fig:efficiency}
\end{figure}

$\mathrm{Trigger}^3$ consists of three main components: CT (Correction Trigger), LT (LLM Trigger), and FT (Fallback Trigger). To explore the impact of different components on the correction performance, we conduct ablation experiments by adding these three components one by one. 
Although CT is the first module that the query goes through during inference, it does not carry out correction and therefore, cannot demonstrate the effect on correction performance. Hence, we add it last.
The base models are the small model and the LLM in a cascade manner.
The ablation results on Commercial and QQ datasets are shown in Table~\ref{tab:ablation}, and we provide detailed discussions for each module below:

$+\mathrm{LT}$: This represents adding the LLM trigger to the base model and integrating LLM. It decides whether to call LLM for specific queries and only calls LLM when necessary. We can observe that adding LT consistently improves performance, reflecting the effectiveness of LT in integrating small models and LLMs.

$+\mathrm{FT}$: This represents adding the fallback trigger, which reviews the correction results. It decides whether to return the original query based on the original and corrected queries. If neither of the models can correct the query, we return the original query. Adding FT improves correction performance on both datasets and all three small models, demonstrating its effectiveness.

\textbf{$+\mathrm{CT}$}: This represents adding the correction trigger, which judges the correctness of the input query. 
For queries that are correct, there is no need for models to correct. Adding CT also improves correction performance. We attribute this improvement to its similar function to FT. Queries that are already correct do not need correction, and having the small model and LLM correct them may actually decrease correction performance.

\subsection{Efficiency Analysis}\label{sec:efficiency}
In the process of deploying the model, considering the possibility of parallel pipeline execution, the portion of the query processed by LLMs often becomes a bottleneck for efficiency. At the same time, a widely recognized basic assumption from previous research~\cite{ramirez2024optimising,lu2023routing} in the field of efficient inference is that smaller models are more inference-efficient than larger models. Based on this concept, similar to~\cite{ding2024hybrid}, we use the proportion of queries addressed by LLM as an indicator to evaluate efficiency, termed as LLM coverage: 
\begin{equation}
\label{eq:llm covergae}
\text{LLM} \; \text{coverage}=\frac{\text{The number of queries corrected by LLM}}{\text{The total number of queries}} \notag
\end{equation}

The average LLM coverage (the mean of the LLM coverage across three small models within the framework) of $\mathrm{Trigger}^3$ and the other three frameworks on two datasets can be found in Figure~\ref{fig:efficiency}.
In conjunction with Table~\ref{tab:main results qwen}, we can find that $\mathrm{Trigger}^3$ maintains high efficiency while improving correction performance, mainly due to the following two reasons:
1) $\mathrm{Trigger}^3$ considers excluding the queries that are correct themselves before making corrections and uses CT to filter out the correct queries.
2) Before $\mathrm{Trigger}^3$ hands over the queries to LLM for correction, it considers that only the queries that LLM can correct are handed over to LLM for processing.

The proportion of queries handled by LLM on three different small models for each framework can be found in Table~\ref{tab:efficency}. Take a concrete example, if the dataset is Commercial, the LLM is Qwen1.5-7B-Chat, and the small model is GECToR, the LLM coverage is 32.09. For about 67.91\% of queries, only a small model is enough. The proportion of queries corrected by other LLMs and small models combinations can be similarly obtained from the examples above. 
We find that $\mathrm{Trigger}^3$ not only maintains high correction performance but also ensures efficiency.

\begin{table}[t]
\label{tab:effiency results qwen} 
\centering
\resizebox{0.95\linewidth}{!}{
\begin{tabular}{l l cc cc}
\toprule 

\multirow{2}{*}{\textbf{Small}} & 
\multirow{2}{*}{\textbf{Framework}}&
\multicolumn{2}{c}{\textbf{Commercial}} & \multicolumn{2}{c}{\textbf{QQ}}                                                           \\  
\cmidrule(lr){3-4}\cmidrule(lr){5-6}

  &&\textbf{F$_{0.5}$}& \textbf{LC}&\textbf{F$_{0.5}$}& \textbf{LC} \\
\midrule
\multirow{4}{*}{GECToR} &Meta Routing & 61.02 & 38.22 &44.02 &39.13\\
&HybridLLM  & 61.72 & 33.77 &44.13&39.82\\
&Margin  & 67.48 & 45.63 &47.11&64.46\\
&$\mathrm{Trigger}^3$  & \textbf{74.60} & \textbf{32.09} &\textbf{57.85}&\textbf{31.24}\\
\cdashline{1-6}
\multirow{4}{*}{BART} &Meta Routing & 68.08 &  16.88&60.73&32.11\\
&HybridLLM & 70.61 & 12.57&61.54&31.02\\
&Margin & 71.35 & 28.69&58.67&33.55\\
&$\mathrm{Trigger}^3$ & \textbf{75.63} & \textbf{3.84} &\textbf{65.27}&\textbf{18.09}\\
\cdashline{1-6}
\multirow{4}{*}{mT5} 
&Meta Routing & 62.67 & 17.45&56.60&26.34\\
&HybridLLM & 64.87 & 21.26 &57.61&26.33\\
&Margin & 65.63 & 12.09&54.57&\textbf{5.97}\\
&$\mathrm{Trigger}^3$ & \textbf{67.25} & \textbf{9.19} &\textbf{59.41}&16.66\\
\bottomrule
\end{tabular}
}
\caption {Efficiency comparisons between $\mathrm{Trigger}^3$ and other frameworks. The boldface indicates optimal performance and optimal efficiency. \textbf{LC} is short for  LLM Coverage, which denotes the proportion of queries solved by LLM. \textbf{F$_{0.5}$} is Char-F$_{0.5}$. Margin is short for Margin Sampling.
}
\label{tab:efficency}
\end{table}

\section{Conclusion}
In this paper, we propose a large-small model collaboration framework, $\mathrm{Trigger}^3$, to adaptively perform query correction. 
Specifically, $\mathrm{Trigger}^3$ uses three triggers to integrate the small model and LLM for query correction. 
First, before performing query correction, it judges the correctness of the query and selects the incorrect query to be corrected by the small model. Second, after the small model correction, it selects the queries that the small model cannot correct but the LLM can, and hands them over to LLM for correction. Finally, after the LLM correction, it reviews and selects the queries that neither the LLM nor the small model can correct, and returns the original query as output.
The superiority and efficiency of $\mathrm{Trigger}^3$'s correction performance are validated through extensive experiments. 

\section*{Acknowledgements}
This work was partially supported by 
the National Natural Science Foundation of China (No. 62376275, 92470205, 62377044), Intelligent Social Governance Interdisciplinary Platform, Major Innovation \& Planning Interdisciplinary Platform for the ``Double-First Class'' Initiative, Renmin University of China. Supported by fund for building world-class universities (disciplines) of Renmin University of China. 
Supported by Public Computing Cloud, Renmin University of China.
Supported by the Fundamental Research Funds for the Central Universities, and the Research Funds of Renmin University of China (23XNKJ13). Supported by Kuaishou Technology.

\bigskip

\bibliography{aaai25}

\clearpage

\appendix

\section{Experimental Settings}

\subsection{More Details on Baselines}\label{appendix:baseline}
The small models consist of the following models:
\begin{itemize}
    \item \textbf{GECToR-Chinese}~\cite{zhang2022mucgec}  is a Seq2Edit model that apapts GECToR~\cite{omelianchuk2020gector} to the Chinese scenario for correction task.
    \item \textbf{BART-Large}~\cite{shao2024cpt} is a Seq2Seq model specially trained for text generation, and we can turn the query correction task into a translatation-like task to complete.
    \item \textbf{mT5-Base}~\cite{xue2021mt5} is another Seq2Seq model, which is based on the T5~\cite{raffel2020exploring} model and pre-trained in multiple languages, so that the model can understand and generate text in multiple languages.
\end{itemize}

The LLMs consist of the following models:
\begin{itemize}
    \item \textbf{Qwen1.5-7B-Chat}~\cite{qwen} is a large language model that performs well in the Chinese field. According to~\cite{fan2023grammargpt}, fine tuning can improve LLM's performance in the field of correction. Here, we use 1K query correction data from training dataset to fine tune the LLM to improve LLM's correction performance.
    \item \textbf{Baichuan2-7B-Chat}~\cite{yang2023baichuan} is another large language model that performs well in the Chinese domain, and we use the same data fine-tuning as above to make it better able to complete the query correction task.
\end{itemize}

\begin{figure}[t]
    \centering
        \includegraphics[width=0.98\linewidth]{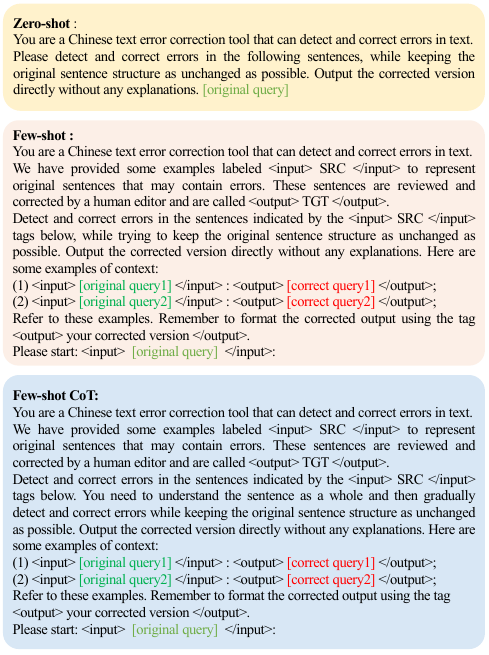}
    \caption{ LLM Templates of Zero-shot, Few-shot and Few-shot CoT.
    }
\label{fig:template}
\end{figure}

The following frameworks are used for comparison:
\begin{itemize}
    \item \textbf{Random-Routing} selects a LLM or a small model for query correction using a random strategy on an incoming query, regardless of other information, as a direct baseline.
    \item \textbf{Routing}~\cite{lu2023routing,vsakota2024fly} trains a model to predict the query correction result of the small model for a query. If the small model fails to achieve the 3 points mentioned in Section~\ref{sec:LT}, it will be handed over to the LLM for correction, without considering the feedback of the LLM.
    \item \textbf{HybridLLM}~\cite{ding2024hybrid}  trains a model to determine whether the smaller model corrects the current query better than the LLM. Better, give it to the small model, otherwise, give it to the LLM.
    \item \textbf{Random-Cascading}, as a direct baseline of the cascading mode, first uses a small model to correct incoming queries, and then uses a random strategy to determine whether LLM is required.
    \item \textbf{Margin Sampling}~\cite{ramirez2024optimising} uses the the small model and the LLM successively in a cascading mode. After the query correction of the small model, the LLM is determined according to the uncertainty of the first token output by the small model.
\end{itemize}

\begin{table*}[!ht]
\centering
\resizebox{\linewidth}{!}{
\begin{tabular}{ll cccccc cccccc}
\toprule 

\multirow{3}{*}{\textbf{Category}}&
\multirow{3}{*}{\textbf{Model}} & \multicolumn{6}{c}{\textbf{Commercial}} & \multicolumn{6}{c}{\textbf{QQ}}                                                                      \\
\cmidrule(lr){3-8}\cmidrule(lr){9-14}&
& \multicolumn{3}{c}{\textbf{Character-level}} & \multicolumn{3}{c}{\textbf{Word-level}}            & \multicolumn{3}{c}{\textbf{Character-level}} & \multicolumn{3}{c}{\textbf{Word-level}}                                                           \\
\cmidrule(lr){3-5}\cmidrule(lr){6-8}\cmidrule(lr){9-11}\cmidrule(lr){12-14}

  && \textbf{P}& \textbf{R}& \textbf{F$_{0.5}$} &  \textbf{P}    &  \textbf{R} & \textbf{F$_{0.5}$} & \textbf{P}& \textbf{R}& \textbf{F$_{0.5}$} &  \textbf{P}    &  \textbf{R} & \textbf{F$_{0.5}$} \\
  
\hline
\multirow{3}{*}{Individual} &
\textbf{GECToR} (Small Model) & 59.59 & \underline{76.30} & 62.32 & 58.68 & \underline{68.71} & 60.44 & 39.96 & 46.10 & 41.05 & 44.59 & 43.69 & 44.41 \\
&Single (LLM) & 43.74 & 42.67 & 43.52 & 44.59 & 40.98 & 43.81 & 36.35 & 40.28 & 37.07 & 38.56 & 36.65 & 38.16 \\
&Cascade (LLM)& \underline{61.36} & 71.26 & \underline{63.12} & \underline{61.25} & 66.47 & \underline{62.23} & 40.25 & \underline{49.94} & 41.88 & 42.69 & \underline{45.80} & 43.28 \\
\cdashline{1-14}
\multirow{6}{*}{Combination} &
Random Routing & 52.58 & 59.50 & 53.83 & 52.30 & 54.74 & 52.77 & 38.66 & 43.62 & 39.56 & 42.12 & 40.54 & 41.80 \\
&Meta Routing & 58.87 & 70.12 & 60.82 & 58.68 & 64.66 & 59.79 & 41.34 & 47.73 & 42.47 & 44.61 & 43.65 & 44.42 \\
&HybridLLM & 59.39 & 73.08 & 61.70 & 59.02 & 66.68 & 60.41 & \underline{41.76} & 47.90 & \underline{42.86} & \underline{45.42} & 44.14 & \underline{45.16}\\
&Random Cascading  & 60.42 & 73.73 & 62.68 & 59.93 & 67.58 & 61.32 & 40.23 & 48.15 & 41.60 & 43.66 & 44.77 & 43.88\\
&Margin Sampling & 60.62 & 73.18 & 62.77 & 60.32 & 67.41 & 61.61 & 40.45 & 47.62 & 41.71 & 44.26 & 44.85 & 44.38\\
&$\mathrm{Trigger}^3$ (Ours) & \textbf{66.86}$^\dagger$ & \textbf{76.42} & \textbf{68.57}$^\dagger$ & \textbf{66.85}$^\dagger$ & \textbf{71.74}$^\dagger$ & \textbf{67.77}$^\dagger$ & \textbf{50.98}$^\dagger$ & \textbf{50.08} & \textbf{50.79}$^\dagger$ & \textbf{54.98}$^\dagger$ & \textbf{47.45}$^\dagger$ & \textbf{53.29}$^\dagger$ \\
\hline
\hline
\multirow{3}{*}{Individual} &
\textbf{BART} (Small Model)  & \underline{73.52} & \underline{71.99} & \underline{73.21} & \underline{73.91} & \underline{71.54} & \underline{73.42} & 59.83 & 60.51 & 59.97 & 62.26 & \underline{62.11} & 62.23 \\
&Single (LLM) & 43.74 & 42.67 & 43.52 & 44.59 & 40.98 & 43.81  & 36.35 & 40.28 & 37.07 & 38.56 & 36.65 & 38.16 \\
&Cascade (LLM)& 63.08 & 64.93 & 63.44 & 63.54 & 63.17 & 63.47 & 45.41 & 56.57 & 47.27 & 48.24 & 53.97 & 49.28 \\
\cdashline{1-14}
\multirow{6}{*}{Combination} &
Random Routing  & 58.86 & 57.46 & 58.57 & 59.81 & 56.34 & 59.08 & 48.17 & 50.78 & 48.67 & 51.29 & 49.65 & 50.96 \\
&Meta Routing  & 70.83 & 68.24 & 70.29 & 71.34 & 67.68 & 70.58 & 57.92 & 60.14 & 58.35 & 61.88 & 60.25 & 61.55 \\
&HybridLLM   & 71.33 & 68.77 & 70.80 & 71.94 & 68.32 & 71.19 & \underline{59.90} & \underline{60.83} & \underline{60.09} & \underline{63.78} & 61.03 & \underline{63.21} \\
&Random Cascading & 68.47 & 68.51 & 68.48 & 68.97 & 67.49 & 68.67 & 52.12 & 58.74 & 53.32 & 55.02 & 58.25 & 55.63 \\
&Margin Sampling & 69.81 & 69.51 & 69.75 & 70.24 & 68.60 & 69.90 & 54.89 & 59.25 & 55.71 & 57.59 & 59.36 & 57.94 \\
&$\mathrm{Trigger}^3$ (Ours) & \textbf{75.03}$^\dagger$ & \textbf{72.19} & \textbf{74.44}$^\dagger$ & \textbf{75.45}$^\dagger$ & \textbf{71.69} & \textbf{74.67}$^\dagger$  & \textbf{63.12}$^\dagger$ & \textbf{61.71}$^\dagger$ & \textbf{62.84}$^\dagger$ & \textbf{66.12}$^\dagger$ & \textbf{62.27}$^\dagger$ & \textbf{65.31}$^\dagger$ \\
\hline
\hline
\multirow{3}{*}{Individual} &
\textbf{mT5} (Small Model)  & \underline{67.42} & 59.04 & \underline{65.56} & \underline{68.44} & 58.02 & \underline{66.06} & 54.71 & 52.01 & 54.15 & 56.61 & 51.02 & 55.40 \\
&Single (LLM) & 43.74 & 42.67 & 43.52 & 44.59 & 40.98 & 43.81 & 36.35 & 40.28 & 37.07 & 38.56 & 36.65 & 38.16 \\
&Cascade (LLM) & 57.92 & 57.15 & 57.77 & 58.73 & 55.37 & 58.03 & 41.75 & 51.58 & 43.41 & 43.71 & 47.54 & 44.43 \\
\cdashline{1-14}
\multirow{6}{*}{Combination} &
Random Routing  & 55.29 & 51.22 & 54.42 & 56.34 & 49.78 & 54.9 & 45.49 & 46.51 & 45.69 & 48.07 & 44.07 & 47.21 \\
&Meta Routing  & 65.23 & 58.51 & 63.77 & 66.33 & 57.48 & 64.35 & 55.73 & 52.00 & 54.94 & 59.15 & 50.42 & 57.17  \\
&HybridLLM  & 67.09 & 58.71 & 65.23 & 68.11 & 57.73 & 65.75 & \underline{58.22} & \underline{52.55} & \underline{56.99} & \underline{61.45} & \underline{51.18} & \underline{59.08} \\
&Random Cascading  & 62.62 & 58.30 & 61.70 & 63.52 & 56.87 & 62.07  & 47.50 & 51.78 & 48.30 & 49.70 & 49.25 & 49.61 \\
&Margin Sampling & 66.77 & \underline{59.56} & 65.19 & 67.77 & \underline{58.40} & 65.67 & 54.17 & 52.16 & 53.75 & 56.08 & 50.98 & 54.98 \\
&$\mathrm{Trigger}^3$ (Ours) & \textbf{68.60}$^\dagger$ & \textbf{59.66} & \textbf{66.61}$^\dagger$ & \textbf{69.56}$^\dagger$ & \textbf{58.62} & \textbf{67.06}$^\dagger$ & \textbf{59.65}$^\dagger$ & \textbf{52.70} & \textbf{58.11}$^\dagger$ & \textbf{62.38}$^\dagger$ & \textbf{51.34} & \textbf{59.81}$^\dagger$ \\
\bottomrule
\end{tabular}
}
\caption {Performance comparisons between $\mathrm{Trigger}^3$ and the baselines when the LLM is Baichuan2-7B-Chat. 
Single: directly using LLM for correction. Cascading: using smaller model rewrites as part of LLM prompts. The LLMs use 1,000 data for fine tuning, while the small model use full training data for training.
The boldface indicates the best performance, and the underline indicates the second performance. `$\dagger$' indicates that the improvements are significant (t-tests, $p\textrm{-value}< 0.05$).
}
\label{tab:main results baichuan} 
\end{table*}

\subsection{More Details on Implementation}\label{appendix:implementation}
For the training of GECToR-Chinese and BART-Large, follwing~\cite{zhang2022mucgec}, we initialize with StructBERT~\cite{wang2019structbert} and Chinese BART-large~\cite{shao2024cpt} respectively. For mT5-Base, we use Mengzi-T5-Base~\cite{zhang2021mengzi} to continue training for the query correction task. 
For the fine tuning dataset, since the small model and the LLM are trained separately, we can know the correction of the small model on the training dataset when we fine-tune LLM. We obtained 1,000 queries in the training dataset, and the correction results from the small model have been obtained. 
Considering the diversity of fine-tuning data, we set half of them for LLM to correct directly, and added preliminary rewriting to the other half.

\begin{table*}[t]
\centering
\resizebox{0.95\linewidth}{!}{
\begin{tabular}{ll cccccc cccccc}
\toprule 

\multirow{3}{*}{\textbf{Category}}&
\multirow{3}{*}{\textbf{Model}} & \multicolumn{6}{c}{\textbf{Commercial}} & \multicolumn{6}{c}{\textbf{QQ}}                                                                      \\
\cmidrule(lr){3-8}\cmidrule(lr){9-14}&
& \multicolumn{3}{c}{\textbf{Character-level}} & \multicolumn{3}{c}{\textbf{Word-level}}            & \multicolumn{3}{c}{\textbf{Character-level}} & \multicolumn{3}{c}{\textbf{Word-level}}                                                       \\
\cmidrule(lr){3-5}\cmidrule(lr){6-8}\cmidrule(lr){9-11}\cmidrule(lr){12-14}

  && \textbf{P}& \textbf{R}& \textbf{F$_{0.5}$} &  \textbf{P}    &  \textbf{R} & \textbf{F$_{0.5}$} & \textbf{P}& \textbf{R}& \textbf{F$_{0.5}$} &  \textbf{P}    &  \textbf{R} & \textbf{F$_{0.5}$} \\
  
\hline
\multirow{3}{*}{Small Model} 
&GECToR  & 59.59 & \textbf{76.30} & 62.32 & 58.68 & 68.71 & 60.44  & 39.96 & 46.10 & 41.05 & 44.59 & 43.69 & 44.41   \\
&BART  & \textbf{73.52} & 71.99 & \textbf{73.21} & \textbf{73.91} & \textbf{71.54} & \textbf{73.42}  & \textbf{59.83} & \textbf{60.51} & \textbf{59.97} & \textbf{62.26} & \textbf{62.11} & \textbf{62.23}   \\
&mT5  & 67.42 & 59.04 & 65.56 & 68.44 & 58.02 & 66.06  & 54.71 & 52.01 & 54.15 & 56.61 & 51.02 & 55.40 \\
\midrule
\multirow{3}{*}{Direct} &
Zero-Shot  & 13.52 & 21.14 & 14.57 & 13.84 & 24.45 & 15.16  & 12.37 & 22.15 & 13.56 & 13.28 & 29.85 & 14.93  \\
&Few-shot & 17.42 & 22.21 & 18.21 & 17.43 & 24.74 & 18.53 & 16.87 & 23.96 & 17.93 & 17.05 & 30.37 & 18.69 \\
&Few-shot-CoT  & 17.67 & 22.77 & 18.50 & 17.67 & 25.41 & 18.81 & 17.31 & 24.89 & 18.43 & 17.49 & 31.50 & 19.19\\
\cdashline{1-14}
\multirow{4}{*}{Fine-Tuning} &
Single  & 45.47 & 42.87 & 44.92 & 45.57 & 40.96 & 44.56   & 41.57 & 40.50 & 41.35 & 43.93 & 37.78 & 42.55 \\
&Cascading (GECToR) & \textbf{72.43} & \textbf{67.13} & \textbf{71.30} & 72.34 & \textbf{64.35} & \textbf{70.59}  & 51.84 & 47.00 & 50.79 & 54.66 & 44.72 & 52.34   \\
&Cascading (BART) & 72.73 & 62.55 & 70.43 & \textbf{73.05} & 61.79 & 70.48 & \textbf{55.73} & \textbf{52.41} & \textbf{55.03} & \textbf{58.57} & \textbf{51.32} & \textbf{56.96}  \\
&Cascading (mT5) & 66.18 & 54.90 & 63.57 & 66.90 & 53.79 & 63.79  & 50.48 & 49.02 & 50.18 & 52.86 & 46.56 & 51.46   \\
\bottomrule
\end{tabular}
}
\caption {Performance comparisons of LLMs when the LLM is Qwen1.5-7B-Chat. 
Single: directly using fine-tuning LLM for correction. Cascading: using smaller model rewrites as part of LLM prompts. 
The boldface indicates the best performance of LLM. 
}
\label{tab:llm issues} 
\end{table*}

\section{Main Results of Baichuan}\label{appendix:baichuan results}
The main results that when the LLM is Baichuan2-7B-Chat are shown in Table~\ref{tab:main results baichuan}.
From the table, we can get conclusions similar to Section~\ref{sec:main results}. On the two query correction datasets and three different small models, $\mathrm{Trigger}^3$ achieves the optimal  correction performance, which also verifies the effectiveness of $\mathrm{Trigger}^3$.

\section{Issues of LLM in Query Correction}\label{appendix llm issues}

According to the conclusions in~\cite{fang2023chatgpt,li2023effectiveness}, LLM has a serious over-correction phenomenon in solving query correction. It will make many unnecessary modifications, which leads to a decline in correction performance. Based on~\cite{fang2023chatgpt}, we directly use LLM for correction on two query correction datasets, and correct through zero-shot, few-shot and few-shot-cot methods. The templates are show in Figure~\ref{fig:template}.

The results are shown in Table~\ref{tab:llm issues}. We can observe that it is difficult to improve the performance of LLM correction just by adjusting the prompt.
Therefore, in this paper, in order to improve the correction performance of LLM in the correction task, the LLMs in the main experiments have been fine-tuned. Also, we add the preliminary rewriting of the small model as an implicit feature in the prompt of LLM to further improve performance.

\end{document}